\documentclass[sigconf]{acmart}

\usepackage{booktabs}
\usepackage{tabularx}
\usepackage{array}
\usepackage{microtype} 
\usepackage{caption} 
\usepackage{placeins} 
\usepackage{cuted}      
\usepackage{dblfloatfix} 
\usepackage{multicol}
\usepackage{enumitem}
\usepackage{placeins}  
\usepackage{ragged2e}  
\usepackage{needspace}

\setlist[itemize]{leftmargin=*, itemsep=2pt, topsep=2pt}
\setlist[enumerate]{leftmargin=*, itemsep=1pt, topsep=2pt}

\usepackage{listings}
\usepackage{xcolor}

\lstdefinelanguage{json}{
  basicstyle=\ttfamily,
  numbers=left,
  numberstyle=\tiny,
  stepnumber=1,
  numbersep=5pt,
  showstringspaces=false,
  breaklines=true,
  frame=lines,
  backgroundcolor=\color{gray!5},
  stringstyle=\color{brown},
  morestring=[b]",
  literate=
   *{0}{{{\color{blue}0}}}{1}
    {1}{{{\color{blue}1}}}{1}
    {2}{{{\color{blue}2}}}{1}
    {3}{{{\color{blue}3}}}{1}
    {4}{{{\color{blue}4}}}{1}
    {5}{{{\color{blue}5}}}{1}
    {6}{{{\color{blue}6}}}{1}
    {7}{{{\color{blue}7}}}{1}
    {8}{{{\color{blue}8}}}{1}
    {9}{{{\color{blue}9}}}{1}
}

\usepackage{enumitem} 
\usepackage{listings}
\usepackage{graphicx}
\usepackage{float}        
\usepackage{placeins}     
\usepackage{dblfloatfix}  
\usepackage{xparse}
\usepackage[font=small,skip=2pt]{caption}
\usepackage{needspace}
\usepackage{multicol}
\usepackage[most]{tcolorbox}
\usepackage{listings}
\usepackage{xcolor} 

\lstdefinestyle{code}{
    language=Python,
    basicstyle=\ttfamily\small,
    breaklines=true,
    keywordstyle=\color{blue},
}

\usepackage{listings}
\usepackage{xcolor}

\definecolor{codegray}{gray}{0.45}
\definecolor{codeblue}{rgb}{0.2,0.35,0.65}
\definecolor{codegreen}{rgb}{0.15,0.45,0.25}
\definecolor{codebg}{rgb}{0.98,0.98,0.98}

\lstdefinestyle{code}{
  backgroundcolor=\color{codebg},
  basicstyle=\ttfamily\small,
  keywordstyle=\color{codeblue}\bfseries,
  stringstyle=\color{codegreen},
  commentstyle=\color{codegray},
  numberstyle=\tiny\color{codegray},
  numbers=left,
  numbersep=8pt,
  frame=single,
  rulecolor=\color{black!15},
  breaklines=true,
  breakatwhitespace=true,
  showstringspaces=false,
  tabsize=2,
  xleftmargin=8pt,
  xrightmargin=2pt,
  keepspaces=true
}

\lstdefinestyle{json}{
  style=code,
  basicstyle=\ttfamily\scriptsize
}

\setcopyright{none}
\acmConference{}{}{}
\copyrightyear{}
\acmISBN{}
\renewcommand\footnotetextcopyrightpermission[1]{}
\NewDocumentCommand{\SectionWideFigure}{m m m}{%
  \FloatBarrier
  \begin{figure*}[!t]
    \centering
    \includegraphics[width=\textwidth]{#1}
    \caption{#2}
    \label{#3}
  \end{figure*}
}

\NewDocumentCommand{\SubsectionFigure}{O{0.32\textheight} m m m}{%
  \begin{figure}[!htbp]
    \centering
    \includegraphics[width=\columnwidth, height=#1, keepaspectratio]{#2}
    \caption{#3}
    \label{#4}
  \end{figure}
  \FloatBarrier
}

\NewDocumentCommand{\HeroFull}{m m m}{%
  \SectionWideFigure{#1}{#2}{#3}
}

\makeatletter
\newcommand{\AppendixOneColumn}{%
  \clearpage
  \onecolumn            
  \setlength{\columnsep}{18pt}
}
\newcommand{\BackToTwoColumn}{%
  \clearpage
  \twocolumn
}
\makeatother

\newcommand{\panel}[1]{\textbf{(#1)}}

\AtBeginDocument{%
  
}

\begin{document}

\title{MineNPC-Task: Task Suite for Memory-Aware Minecraft Agents}

\author{Tamil Sudaravan Mohan Doss}
\affiliation{%
  \institution{Microsoft}
  \city{}
  \country{}}
\email{tsudaravanm@microsoft.com}

\author{Michael Xu}
\affiliation{%
  \institution{Microsoft Research}
  \city{}
  \country{United States}}
\email{michaelxu@microsoft.com}

\author{Sudha Rao}
\affiliation{%
  \institution{Microsoft Research}
  \city{}
  \country{United States}}
\email{Sudha.Rao@microsoft.com}

\author{Andrew D. Wilson}
\affiliation{%
  \institution{Microsoft Research}
  \city{}
  \country{United States}}
\email{awilson@microsoft.com}

\author{Balasaravanan Thoravi Kumaravel}
\affiliation{%
  \institution{Microsoft Research}
  \city{}
  \country{United States}}
\email{bala.kumaravel@microsoft.com}

\renewcommand{\shortauthors}{Mohan Doss et al.}

\newcommand{\benchmarkname}{\textsc{MineNPC-Task}}
\newcommand{\systemnamedot}{\texttt{Thor }}  
\newcommand{\systemname}{\texttt{~Thor}}     
\newcommand{\etal}{\textit{et al.}}
\newcommand{\eg}{e.g., }
\newcommand{\ie}{i.e., }


\newcommand{\tamil}[1]{}
\newcommand{\andy}[1]{}
\newcommand{\bala}[1]{}
\newcommand{\sudha}[1]{}
\newcommand{\michael}[1]{}

\begin{abstract}
We present \textsc{MineNPC-Task}, a user-authored benchmark and evaluation harness for testing memory-aware, mixed-initiative LLM agents in open-world \emph{Minecraft}. Rather than relying on synthetic prompts, tasks are elicited from formative and summative co-play with expert players, normalized into parametric templates with explicit preconditions and dependency structure, and paired with machine-checkable validators under a bounded-knowledge policy that forbids out-of-world shortcuts. The harness captures plan/act/memory events-including plan previews, targeted clarifications, memory reads and writes, precondition checks, and repair attempts and reports outcomes relative to the total number of attempted subtasks, derived from in-world evidence.

As an initial snapshot, we instantiate the framework with GPT-4o and evaluate \textbf{216} subtasks across \textbf{8} experienced players. We observe recurring breakdown patterns in code execution, inventory/tool handling, referencing, and navigation, alongside recoveries supported by mixed-initiative clarifications and lightweight memory. Participants rated interaction quality and interface usability positively, while highlighting the need for stronger memory persistence across tasks. We release the complete task suite, validators, logs, and harness to support transparent, reproducible evaluation of future memory-aware embodied agents.
\end{abstract}

\begin{teaserfigure}
  \centering
  \includegraphics[width=\textwidth]{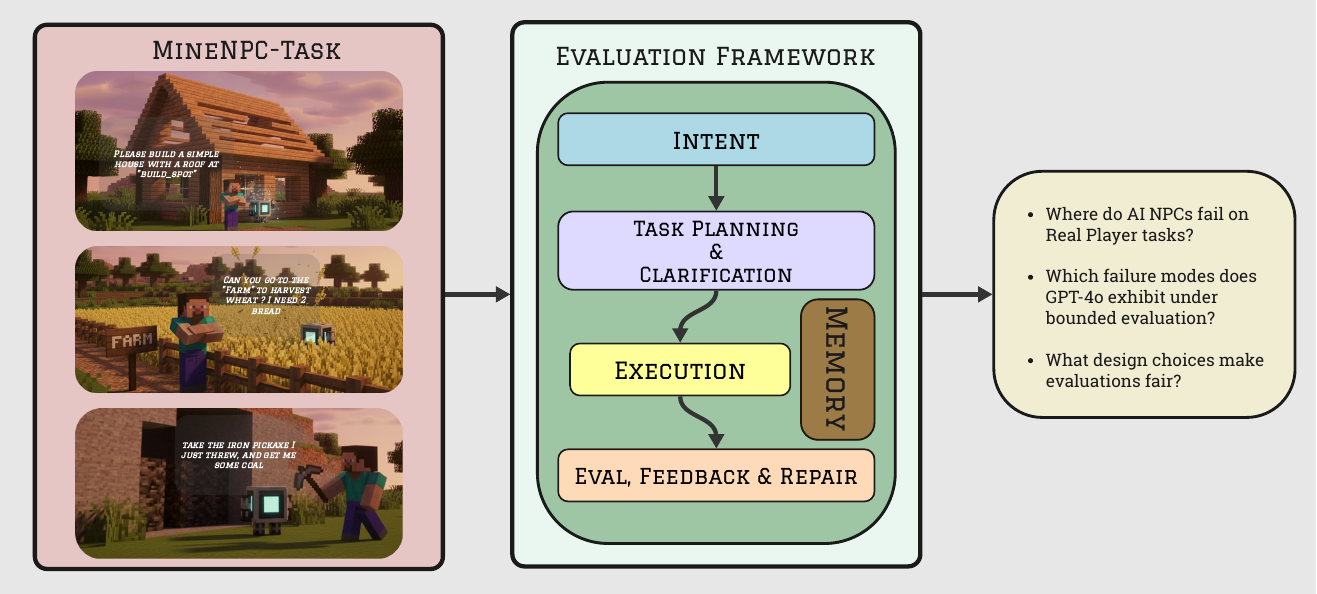}
  \caption{\textbf{}
  \panel{a} \textsc{MineNPC-Task}: Each row depicts a naturalistic, player-authored request in Minecraft, grounding evaluation in distinct phases of co-play building, farming, and mining.
  \panel{b} \textbf{Bounded, reproducible evaluation}: a model-agnostic harness that routes intents, plans with single-turn clarification, executes via public Mineflayer APIs under a bounded-knowledge policy, and judges only from in-world evidence with lightweight memory.
  \panel{c} \textbf{Questions this benchmark enables}: Where do AI NPCs fail on real player tasks? Which failure modes does \mbox{GPT-4o} exhibit under this bounded setup? Which design choices make evaluations fair?}
  \Description{Left: A vertical stack of Minecraft scenes, each showing one natural player request—build house, harvest wheat, mine coal. Center: block diagram of the evaluation framework with intent, plan, execute, and memory modules. Right: a callout listing three benchmark questions.}
  \label{fig:teaser}
\end{teaserfigure}


\maketitle

\section{Introduction}

Building truly capable AI companions for open-world games requires more than one-shot instruction following, it requires agents that can \emph{plan}, \emph{clarify}, and \emph{remember} in order to operate effectively in dynamic, long-horizon environments. However, existing evaluation practices fall short: many rely on overly prescriptive prompts or grant agents privileged access to hidden environment state, artificially inflating their apparent competence and undermining fair comparison across models. This highlights the need for a \emph{transparent, reproducible benchmark} that captures the pressures of mixed-initiative interaction~\cite{Lai2022MixedInitiative}, without granting unfair advantages. Recent efforts, such as the Minecraft Universe (MCU) benchmark~\cite{MCU2023,MCU2023arxiv}, move in this direction by introducing scalable, composable tasks in open-world environments and emphasizing repeatable, human-aligned evaluation.

We introduce \textsc{MineNPC-Task}, a practical benchmark for evaluating memory-aware, mixed-initiative LLM agents in \emph{Minecraft}. Rather than synthetic prompts, tasks are elicited from expert co-play: during our formative and evaluation sessions, experienced players issued real requests, which we normalized into compact templates with explicit preconditions and a small set of slot parameters, and paired with simple machine-checkable validators. The benchmark runs inside a Mineflayer envelope so perception and action are limited to public, in-game APIs; a bounded-knowledge policy forbids admin commands, global map introspection, and scans beyond loaded chunks. The aim is straightforward: a clean, reusable setup that others can run, swap in different models, and obtain comparable numbers, while avoiding hidden shortcuts that have complicated evaluation in prior \emph{Minecraft}-based agents \citep{johnson2016the,fan2022minedojo,wang2023voyager}.

To execute tasks reproducibly we use a model-agnostic evaluation framework (Section~\ref{sec:framework}). The agent presents a brief plan preview that breaks the request into a handful of subtasks, asks a targeted clarifying question only when a slot is unbound (for example, a tool variant or a search radius), acts through Mineflayer skills, and is judged solely on in-world evidence drawn from inventory and equipment deltas, position changes, nearby entities and blocks within loaded chunks, and recent chat. This mixed-initiative pattern aligns with recent agent designs that combine planning with lightweight memory and reflection while keeping behavior legible to human partners \citep{10.1145/3586183.3606763, sarch2024vlm, Hou_2024}. Short scenario walkthroughs (Section~\ref{sec:scenario-walkthroughs}) illustrate how planning, clarification, and reuse of prior context work together.

As an initial snapshot, we instantiate the framework with GPT-4o and run live co-play with \textbf{8} experienced players (Section~\ref{sec:evaluation}). Across \textbf{44} user-authored tasks and \textbf{216} subtasks, we observe \textbf{71} subtask failures (approximately \textbf{33\%}) and report where things went wrong, including code execution, inventory handling, referencing, and navigation, alongside common recoveries such as clarifying a slot, simplifying a goal, or constraining location. Participants rated interaction quality and interface usability positively, with mixed views on personalization and completion, consistent with the need for stronger memory scaffolding. We intentionally omit ablations or model comparisons. The intent is to provide a compact, user-authored benchmark that others can extend. For context, our focus complements broader surveys on AI in games and adaptive interaction \citep{yannakakis2018artificial, Riedl2021} and ongoing efforts to evaluate reasoning in live games \citep{hu2024gamearena}.

The following are the contribution of this work:
\begin{itemize}[leftmargin=*, itemsep=2pt, topsep=2pt]
  \item \textbf{\textsc{MineNPC-Task} suite:} a user-authored benchmark for \emph{Minecraft}, normalized into templates with explicit preconditions and paired with simple, machine-checkable validators under a bounded-knowledge policy. See Appendix~\ref{app:tasksuite} and Section~\ref{sec:setting}. Compared with prior \emph{Minecraft} agents and datasets, we emphasize human-elicited goals, public-API constraints, and validator-backed judging \citep{johnson2016the,fan2022minedojo,wang2023voyager}.
  \item \textbf{Evaluation framework:} a model-agnostic procedure that enforces plan previews and single-turn clarifications when required, constrains perception and action to Mineflayer APIs, and produces validator-backed outcomes suitable for controlled comparisons across LLMs. See Section~\ref{sec:framework}. The design follows calls for transparent, reproducible evaluation in interactive systems \citep{Jennings2024,hu2024gamearena}.
  \item \textbf{Empirical snapshot with GPT-4o:} co-play results from \textbf{8} experts over \textbf{44} tasks, comprising \textbf{216} subtasks with \textbf{71} failures (approximately \textbf{33\%}), along with observations about where mixed-initiative interaction and lightweight memory help—and where brittleness remains. See Section~\ref{sec:evaluation}.
\end{itemize}
\HeroFull{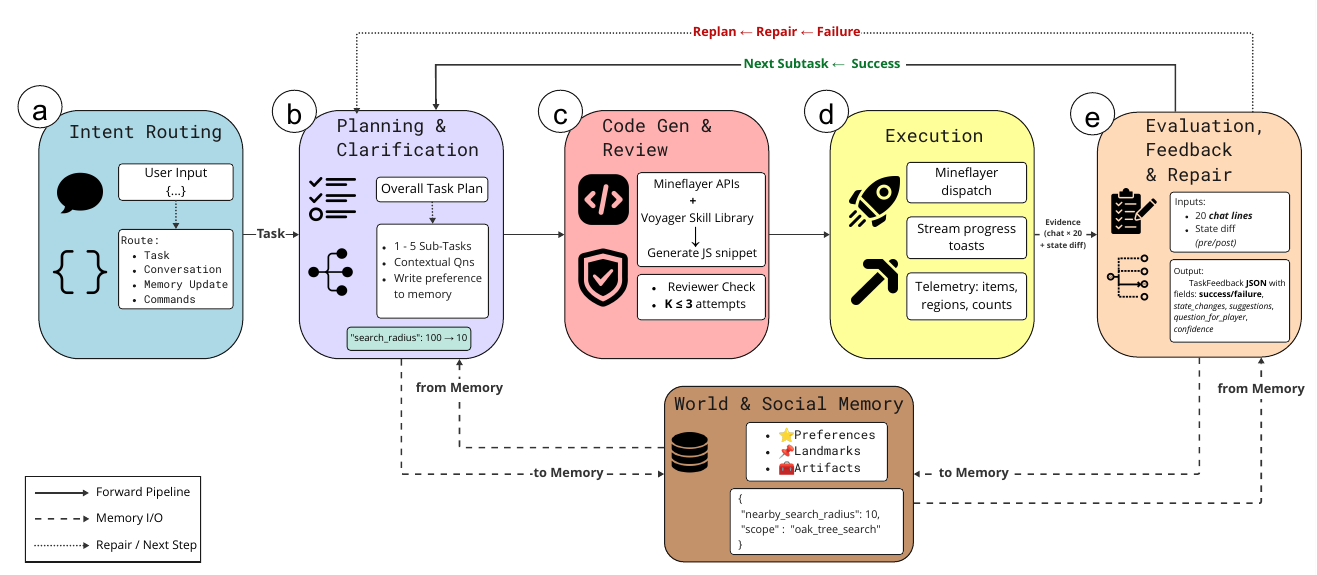}
{\textbf{Plan-Clarify-Act-Judge: our model-agnostic evaluation framework.}
\panel{a} \emph{Intent routing} parses chat into \{\texttt{intent}, \texttt{slots}, \texttt{confidence}\}.
\panel{b} \emph{Planning and clarification} compiles a short plan (3–5 steps); if a required slot is missing, the agent issues a single, contextual question.
\panel{c} \emph{Code generation and review} synthesizes a small JavaScript snippet against Mineflayer APIs and a skill library; a lightweight reviewer caps retries ($K{\le}3$).
\panel{d} \emph{Execution} dispatches approved code and streams concise progress updates.
\panel{e} \emph{Evaluation and bounded repair} reads recent chat and state deltas to emit \texttt{TaskFeedback}; on success the harness advances to the next subtask, and on failure it offers a bounded repair and partial replan. Dashed arrows denote reads/writes to \emph{memory} (landmarks, artifacts, preferences, commitments).}
{fig:atpes-hero}

\section{Related Work}

\noindent\textbf{Benchmarks and evaluation for embodied agents.}
A long line of embodied benchmarks has standardized APIs, sensors, and success criteria for instruction following, navigation, and manipulation. Suites that span vision, language, and action such as ALFRED \citep{shridhar2020alfred}, TEACh \cite{teach2021}, and EmbodiedQA \citep{das2018embodiedqa} probe multistep execution under natural language, often requiring dialog to bind underspecified slots. Platform efforts including Habitat, iTHOR, and ProcTHOR provide large scale scene generation and reproducible perception and action loops \citep{savva2019habitat,ithor,procthor2022}, while BEHAVIOR and VirtualHome emphasize program like decompositions of everyday activities and structured evaluation protocols \citep{behavior2021,puig2018virtualhome}. Text centric environments and curricula such as TextWorld, ALFWorld, ScienceWorld, and BabyAI offer controllable abstractions for reasoning, exploration, and sample efficiency \citep{textworld2018,alfworld2021,scienceworld2022,babyai2019}. Beyond simulators, LLM agent evaluations study open ended tool use and web interaction (AgentBench, WebArena) \citep{agentbench2023,webarena2023}, and analyze reasoning in live games and across modalities \citep{hu2024gamearena,momente2025triangulating,yannakakis2018artificial}. Collectively, these efforts advanced reproducibility and coverage, while differing in reliance on scripted versus user authored goals and on privileged versus in situ evidence for judging.

\noindent\textbf{LLM agents and NPCs in games}
\emph{Minecraft} has served as a versatile laboratory for grounded learning, planning, and human–AI teaming. Malmo established a widely used interface and experimental substrate \citep{johnson2016the,hofmann2019minecraft}. LLM-driven systems leverage internet-scale priors for broad competence: MineDojo aggregates cross-modal knowledge and goals \citep{fan2022minedojo}; Voyager demonstrates autonomous skill acquisition and reuse in open-ended play \citep{wang2023voyager}; STEVE-1 targets text-to-behavior generation \citep{lifshitz2024steve1generativemodeltexttobehavior}; and Ghost in the Minecraft augments agents with text-based knowledge and memory \citep{zhu2023ghostminecraftgenerallycapable}. Parallel threads examine runtime code generation for gameplay and NPC behaviors \citep{Jennings2024,volum-etal-2022-craft} and conversational NPCs that assist with scripted quest structures \citep{rao2024minecraftQuest}. Outside \emph{Minecraft}, game-based evaluations continue to investigate agent reasoning, adaptation, and spectator-facing dynamics across genres and tasks \citep{cheung2011starcraft,yannakakis2018artificial}.

\noindent\textbf{Mixed-initiative planning, clarification, and memory.}
Mixed-initiative pipelines and prompting strategies interleave thinking with acting, emphasizing targeted questions when parameters are unbound and short, legible plans. Dialog-driven instruction following foregrounds clarification for slot binding (e.g., TEACh) \citep{teach2021}, and prompting methods such as ReAct integrate reasoning traces with tool calls \citep{yaoproreact2023}. Foundational agent models formalize state, intentions, and cooperation \citep{Wooldridge_Jennings_1995}, while cognitive distinctions between episodic and semantic memory inform how agents might store and retrieve context \citep{tulving1983elements,tulving1985memory}. Recent LLM systems investigate experience distillation and dynamic memory consolidation for situated tasks \citep{sarch2024vlm, Hou_2024}. Interaction-design perspectives highlight dynamic grounding and constructive negotiation for aligning human–agent work \citep{vaithilingam2024imagining}, and studies of theory-of-mind cues examine coordination in multi-agent settings \citep{li2023theory}. Complementary literatures on NPC believability, social presence, and anthropomorphism provide additional frames for evaluating expectations in game worlds \citep{articleonnpdbeilievability, Nowak2003}.

\section[Benchmark Setting and \textsc{MineNPC-Task}]{Benchmark Setting and \textsc{MineNPC-Task}}
\label{sec:setting}

We instantiate open-world co-play in \emph{Minecraft} using a Mineflayer client so that actions are constrained to public, in-game interfaces. Concretely, the agent observes chat, its own inventory and equipment, and nearby entities/blocks within currently loaded chunks (a practical proxy for line of sight). Actions are issued through high-level skills like \emph{navigate, mine, craft, place, interact} implemented atop Mineflayer APIs. A bounded-knowledge policy forbids privileged capabilities (e.g., \texttt{/give}, \texttt{/teleport}), global map/seed introspection, and scans beyond loaded chunks; runs that violate this policy are invalidated.

\textbf{Task source and structure.} The \textsc{MineNPC-Task} suite is derived from goals observed during expert co-play rather than from synthetic prompts. From these observations we specify a lightweight \emph{task template} that the evaluation framework instantiates at run time. Given a user goal, the framework compiles a short plan (typically 3–5 subtasks) and represents each subtask as a compact record with five fields: name, dependencies, required parameters, a single targeted clarifying question issued only if a required parameter is missing, and a success criterion. Completion is judged by simple, machine-checkable validators that consume only bounded, in-world evidence—inventory/equipment and position deltas, nearby entities/blocks within loaded chunks, and a short window of recent chat—and return a pass/fail with a brief rationale. (Prompts and template examples appear in the Appendix.)

\textbf{Coverage.} The \textsc{MineNPC-Task} suite covers the everyday goals we observed most frequently: \emph{resource acquisition and logistics} (gather–craft chains, inventory constraints); \emph{navigation and retrieval} (recalling named landmarks and fetching/returning items); \emph{tooling and crafting} (multi-step recipes, tool selection with durability checks); \emph{construction} (layout-constrained builds using existing facilities); \emph{combat and safety} (simple loadout preconditions); and \emph{continuity/mixed-initiative} cases where an underspecified slot is bound via a brief clarification. Section~\ref{sec:framework} details how the evaluation framework enforces this template end-to-end and produces validator-backed outcomes.
\begin{figure*}[t]
  \centering
  \includegraphics[width=\textwidth]{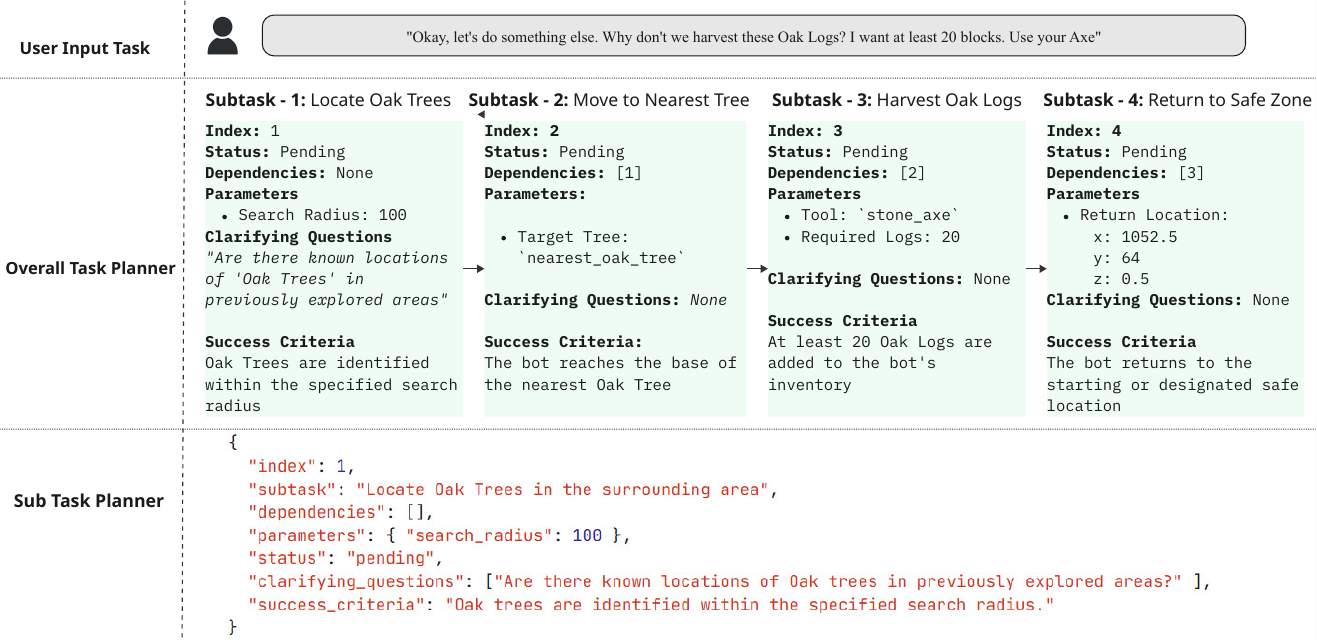}
  \caption{Planning surfaces for “harvest oak logs.”
  \panel{a} Chat request routed to \texttt{task(request)}.
  \panel{b} Short, legible plan with dependencies.
  \panel{c} Subtask record with defaults (e.g., \texttt{search\_radius}=100) and its clarifying question.}
  \label{fig:planner}
\end{figure*}

\section{Evaluation Framework}
\label{sec:framework}

We now describe the evaluation framework that turns the constraints and task templates into a reproducible harness for testing LLM agents in \emph{Minecraft} via Mineflayer. Figure~\ref{fig:atpes-hero} summarizes the pipeline. The framework is model-agnostic: an LLM proposes a short plan, asks at most one targeted clarifying question when a required parameter is unbound, generates Mineflayer-compatible code to act, and is judged from bounded, in-world evidence. In this paper we instantiate the framework with GPT-4o; Section~\ref{sec:evaluation} reports that snapshot.

\SubsectionFigure{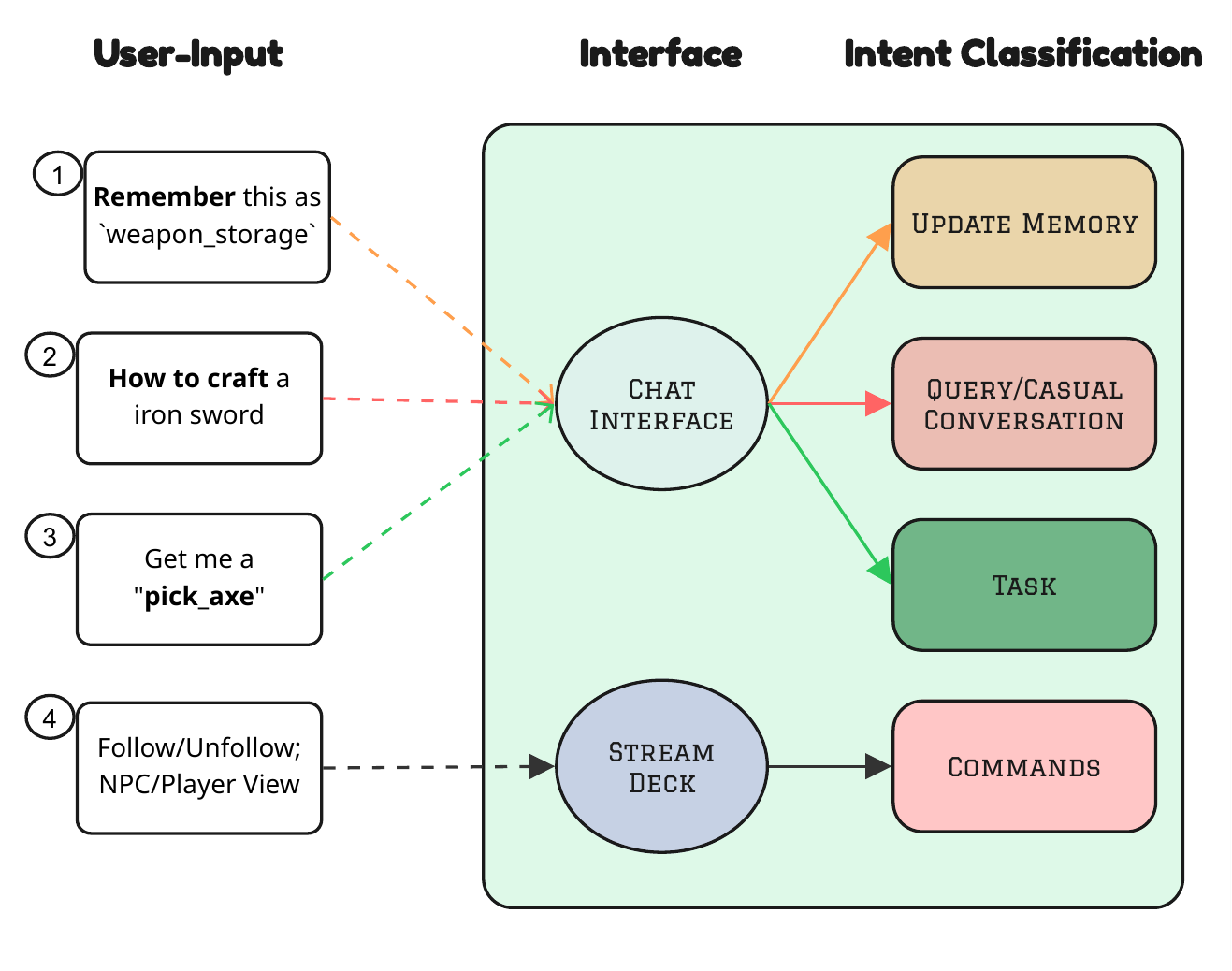}
{Routing and planning. Ingress via chat or control buttons is parsed into intents (\emph{update memory}, \emph{conversation}, \emph{task}, \emph{control}). Tasks flow to a planner that decomposes requests and binds missing slots with a single, contextual question when needed.}
{fig:user-intent}

\noindent
\textbf{Observation and action contract.}
Perception and control are constrained to public, in-game interfaces exposed by Mineflayer. The agent can read recent chat, its inventory and equipment, and nearby entities and blocks within currently loaded chunks (a practical proxy for line-of-sight). It can navigate, mine, craft, place, interact, transfer items, and return/drop off; destructive operations require an explicit confirmation step. A bounded-knowledge policy forbids admin commands (e.g., \texttt{/give}, \texttt{/teleport}), global map/seed introspection, and bulk scans beyond loaded chunks; attempts that trigger a forbidden call are marked invalid. See Fig.~\ref{fig:user-intent} for the routing pathway that feeds planning.

\begin{center}
  \includegraphics[width=\columnwidth]{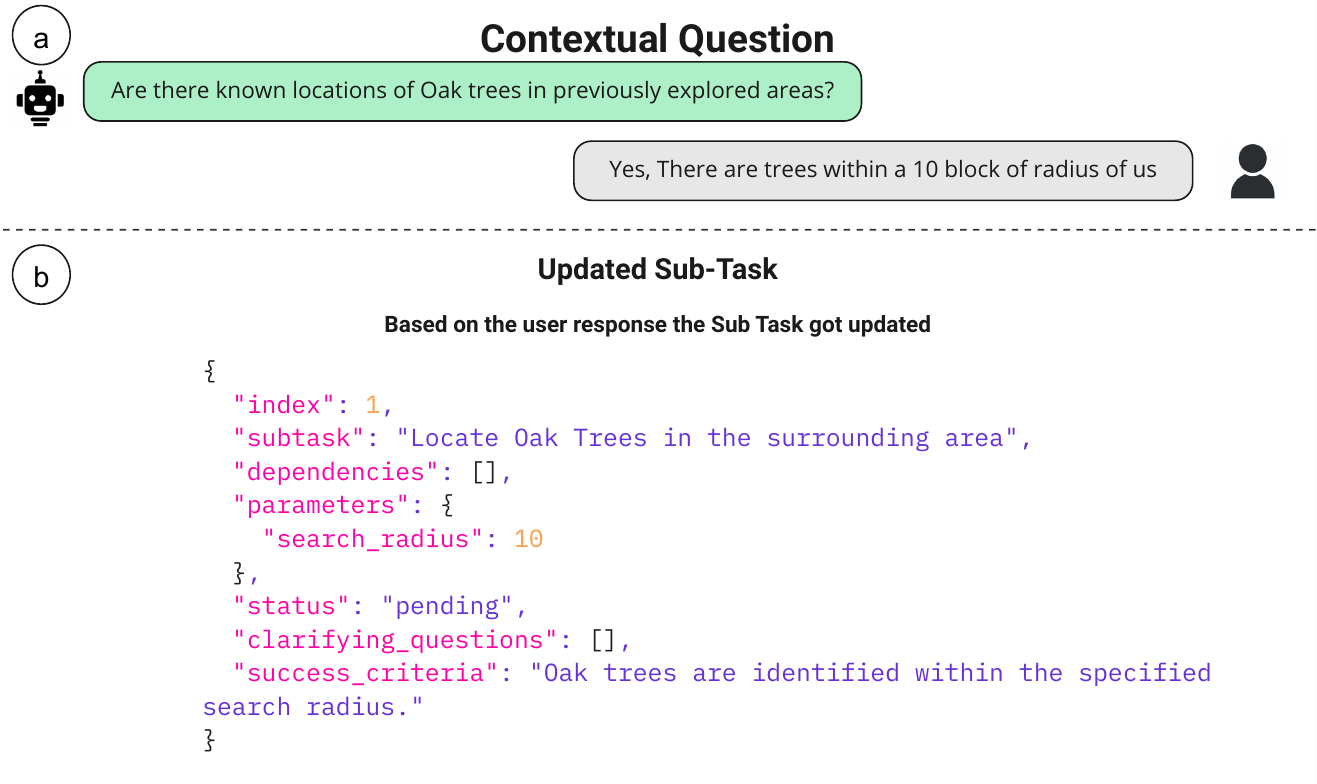}
  \captionof{figure}{Mixed-initiative example. \panel{a} The planner asks whether oak trees are within a known radius; the player replies “within 10 blocks.” \panel{b} The plan updates \texttt{search\_radius} from 100 to 10 and stores the preference with provenance \textit{told}.}
  \label{fig:context-qn}
\end{center}

\noindent
\textbf{Plan preview and clarification.}
Player requests arrive through chat and are routed to \texttt{task(request)} when they imply action. Before any execution, the framework renders a one-line plan preview and exposes precondition checks (Fig.~\ref{fig:planner}). If a required slot (e.g., tool variant, drop-off location, or search radius) is not bound by context, the agent asks \emph{one} targeted question rather than guessing; the answer is applied immediately to the active plan and written to memory with provenance for later reuse (Fig.~\ref{fig:context-qn}).

\noindent
\textbf{From template to plan.}
Given a user goal and the corresponding task template (Appendix~\ref{app:tasksuite}), the framework compiles a short plan with three to five subtasks (Fig.~\ref{fig:planner}\panel{b}). Each subtask is represented by a compact record (Fig.~\ref{fig:planner}\panel{c}) specifying its name, dependencies, required parameters, a clarifying question issued only when a required parameter is missing, and a success criterion. These structures are logged to make plan deltas auditable (e.g., \texttt{search\_radius}: 100$\rightarrow$10).

\noindent
\textbf{Execution with review.}
Plans execute through generated JavaScript against Mineflayer. A lightweight reviewer checks basic API usage and guard conditions; retries are capped ($K{=}2$–$3$) to prevent runaway loops. Approved code is dispatched, and the runtime emits compact progress toasts so players can track what is happening without breaking flow (Fig.~\ref{fig:code-review}).

\begin{center}
  \includegraphics[width=1.1\columnwidth]{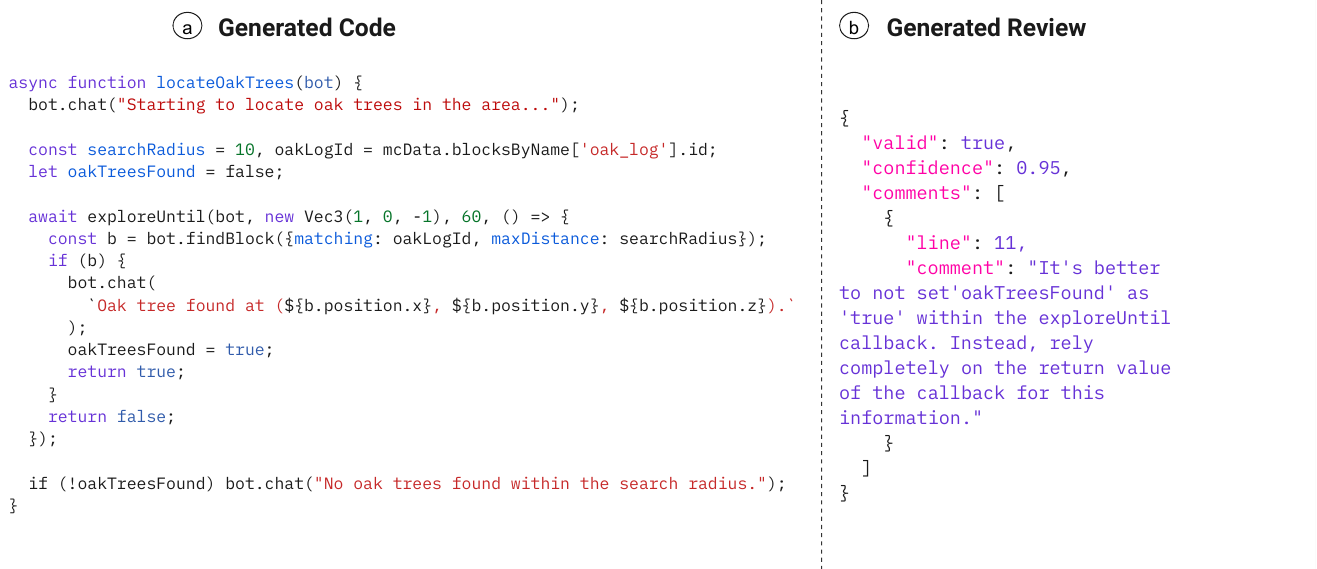}
  \captionof{figure}{Code generation and review. \panel{a} Generated JavaScript reflects updated parameters (e.g., \texttt{searchRadius = 10}) and emits legible chat updates. \panel{b} Reviewer feedback flags robustness issues before execution; retries are capped.}
  \label{fig:code-review}
\end{center}

\noindent\textbf{Judging and bounded repair.}
Completion is judged only from signals available inside the game: pre/post snapshots of inventory, equipment, and position; the presence of nearby entities or blocks within currently loaded chunks; and a short window of recent chat. The evaluator produces a structured \texttt{TaskFeedback} record that states success or failure, cites the state changes it relied on, and may include a brief suggestion or a targeted follow-up question when additional context is required (Fig.~\ref{fig:taskfeedback}). On success, the next subtask begins. On failure or low confidence, the framework presents a small, bounded repair prompt and \emph{the player chooses what happens next}: for example, retry with an adjusted parameter, backtrack to a prior substep, or pause for guidance. If the player issues a new or revised goal, the router treats it as a fresh \texttt{task(request)} and compiles a new plan; otherwise the harness performs a partial replan from the failing step using the player’s selection. This keeps the loop explicitly mixed-initiative: the agent diagnoses and proposes, the player decides, and both the decision and plan deltas are logged.

\medskip
\noindent
\textbf{Memory.}
A simple typed store persists named landmarks, artifacts, preferences, and records of commitments/breakdowns. Each entry carries provenance (\textit{seen}, \textit{told}, \textit{inferred}) and can be retrieved with nearest-$k$ queries scoped to the current task. Answers to clarifying questions are written as scoped preferences that seed future slot values when context is similar.

\medskip
\noindent
\textbf{Reproducibility and limits.}
For each request we log routing latency, plan build time, whether a clarifying question was issued and answered, plan deltas, code-generation and review iterations, execution time, retries and repairs issued/accepted, success/attempt denominators, memory reads/writes (including preference hits/misses), and token usage per stage. Prompts, planner templates, validator stubs, and adapter scaffolding are provided to support replication. Two practical limits are visible in logs: conversational turns can be misrouted as tasks at low confidence (we fall back to a compact keyword heuristic), and scoped preferences can go stale when the world changes off-screen (we mark such entries “stale” and reconfirm before reuse).

\medskip
\noindent
Together, the components above define a complete execution loop for the agent. The next section demonstrates how this framework operates in practice through scenario walkthroughs. These examples focus on typical interactions and highlight how planning, clarification, execution, and judgment unfold step by step.


\begin{strip}
  \centering
  \includegraphics[width=1.0\textwidth]{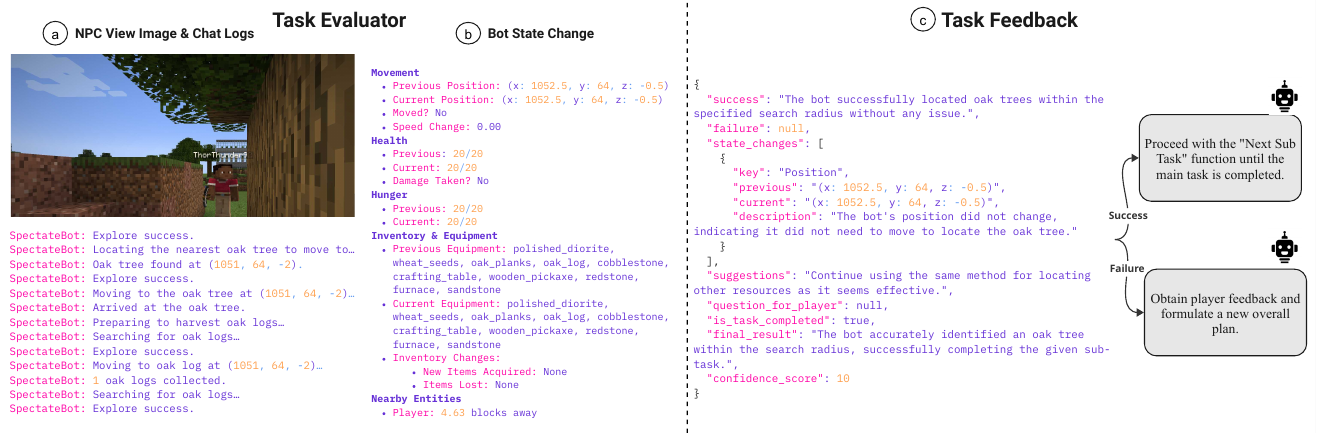}
  \captionof{figure}{Judging and repair. \panel{a} Evaluator consumes bounded evidence: a short window of recent chat and an NPC camera frame. \panel{b} State deltas between pre/post snapshots (position, inventory/equipment, nearby entities/blocks). \panel{c} Structured \texttt{TaskFeedback} with success/failure and rationale, state changes, suggestions, optional question for the player, and a confidence score.}
  \label{fig:taskfeedback}
\end{strip}

\FloatBarrier
\Needspace{6\baselineskip}
\section{Scenario Walkthroughs of \systemname{}}
\label{sec:scenario-walkthroughs}

\noindent
These walkthroughs show how the system is intended to behave when the framework and model work well together. They illustrate plan previews, brief clarifying questions, in-game execution through Mineflayer, and outcome judgment by the validator. The walkthroughs are meant to be illustrative, not evaluative; empirical results, including failures, are reported in Section~\ref{sec:evaluation}.


\vspace{-2.5mm}

\subsection{S1: Bookmark a landmark, then fetch a tool}

\noindent
\textbf{Memory write.}
During casual exploration, the player types:
\emph{“remember this as \texttt{weapon\_storage}.”}
The utterance is routed as \texttt{memory\_update} (Fig.~\ref{fig:user-intent}) and stored as a named landmark with coordinates and provenance \textit{told}. The interface remains legible throughout. Primary play stays in the main window, while a small companion camera indicates the agent’s current view (Fig.~\ref{fig:ui-appx}).

\noindent
\textbf{Request and plan preview.}
Later, at a mine entrance, the player issues:
\emph{“get me a \texttt{pick\_axe}.”}
The router classifies this as \texttt{task(request)}, after which the planner compiles a short plan and renders a one-line preview (Fig.~\ref{fig:planner}):
\emph{go to \texttt{weapon\_storage} $\rightarrow$ select a pickaxe $\rightarrow$ return}.
The active subtask exposes its parameters and defaults for inspection (Fig.~\ref{fig:planner}\panel{c}).

\noindent
\textbf{Single clarification.}
When multiple pickaxes satisfy the request, the planner asks exactly one targeted question (Fig.~\ref{fig:user-intent}):
\emph{“Which pickaxe, iron or diamond?”}
The response is applied immediately to the active subtask and persisted as a scoped preference with provenance \textit{told}.

\noindent
\textbf{Execution and judgment.}
Mineflayer code is generated, reviewed, and executed with capped retries (Fig.~\ref{fig:code-review}). Completion is judged using bounded in-world evidence, including state deltas, nearby entities or blocks, and a short window of recent chat (Fig.~\ref{fig:taskfeedback}). On success, the plan advances. Otherwise, a bounded repair prompt allows the player to choose how to proceed.

\noindent\textit{Logged.}
Routing and plan-build latency; clarification issued and answered; memory reads and writes; code-generation and review iterations; execution time; validator decision and rationale.

\vspace{-2.5mm}
\subsection{S2: Collect 20 oak logs}

\noindent
\textbf{Request and defaults.}
The player requests:
\emph{“collect 20 oak logs.”}
The planner previews a compact plan in chat (Fig.~\ref{fig:planner}):
\emph{locate trees $\rightarrow$ move to nearest tree $\rightarrow$ harvest to \texttt{count=20} $\rightarrow$ return}.
With no relevant prior context, the first subtask adopts a conservative default \texttt{search\_radius} of 100 (Fig.~\ref{fig:planner}\panel{c}).

\noindent
\textbf{Contextual update.}
Before acting, the planner asks a single contextual question (Fig.~\ref{fig:context-qn}\panel{a}):
\emph{“Are there known oak trees nearby?”}
The reply,
\emph{“within 10 blocks,”}
updates the active plan in place (\texttt{search\_radius} \(100 \rightarrow 10\)) and is written as a scoped preference for future oak-tree searches (Fig.~\ref{fig:context-qn}\panel{b}).

\noindent
\textbf{Execution and judgment.}
Generated code reflects the updated parameter and emits legible progress updates during execution (Fig.~\ref{fig:code-review}). A lightweight review flags robustness issues before dispatch. The evaluator produces a structured \texttt{TaskFeedback} record from bounded evidence (Fig.~\ref{fig:taskfeedback}). On success, execution proceeds to the next subtask. On failure or low confidence, the player selects a bounded repair such as retrying with an adjusted parameter, backtracking, or revising the goal.

\noindent\textit{Logged.}
We record plan preview latency, whether a clarification was issued and answered, exact plan deltas, distance to first harvest, time to completion, memory preference writes, and the validator’s decision and rationale.

\noindent
Together, these scenarios capture the benchmark’s intended interaction loop: \emph{preview} a short plan, \emph{bind} missing slots with a single targeted question, \emph{execute} within Mineflayer’s envelope, and \emph{judge} outcomes from bounded in-world evidence, while keeping the human explicitly in the loop for recovery when execution does not go as planned.

\newcommand{\NumParticipants}{8}
\newcommand{\SubtaskFailureRate}{71/216\,$\approx$\,33\%}
\vspace{-7pt}

\section{Evaluating the Framework with GPT-4o and Expert Players}
\label{sec:evaluation}

Having established the harness, validators, memory modules, and UX pipeline, we instantiated the complete framework with GPT-4o and conducted real-time co-play sessions with experienced Minecraft players. The evaluation goal is intentionally modest: characterize how the system behaves end-to-end on user-authored tasks under bounded, reproducible judging—\emph{no ablations, no cross-model comparisons, and no speculative claims}.

\begin{figure}[ht]
  \centering
  \includegraphics[width=\columnwidth]{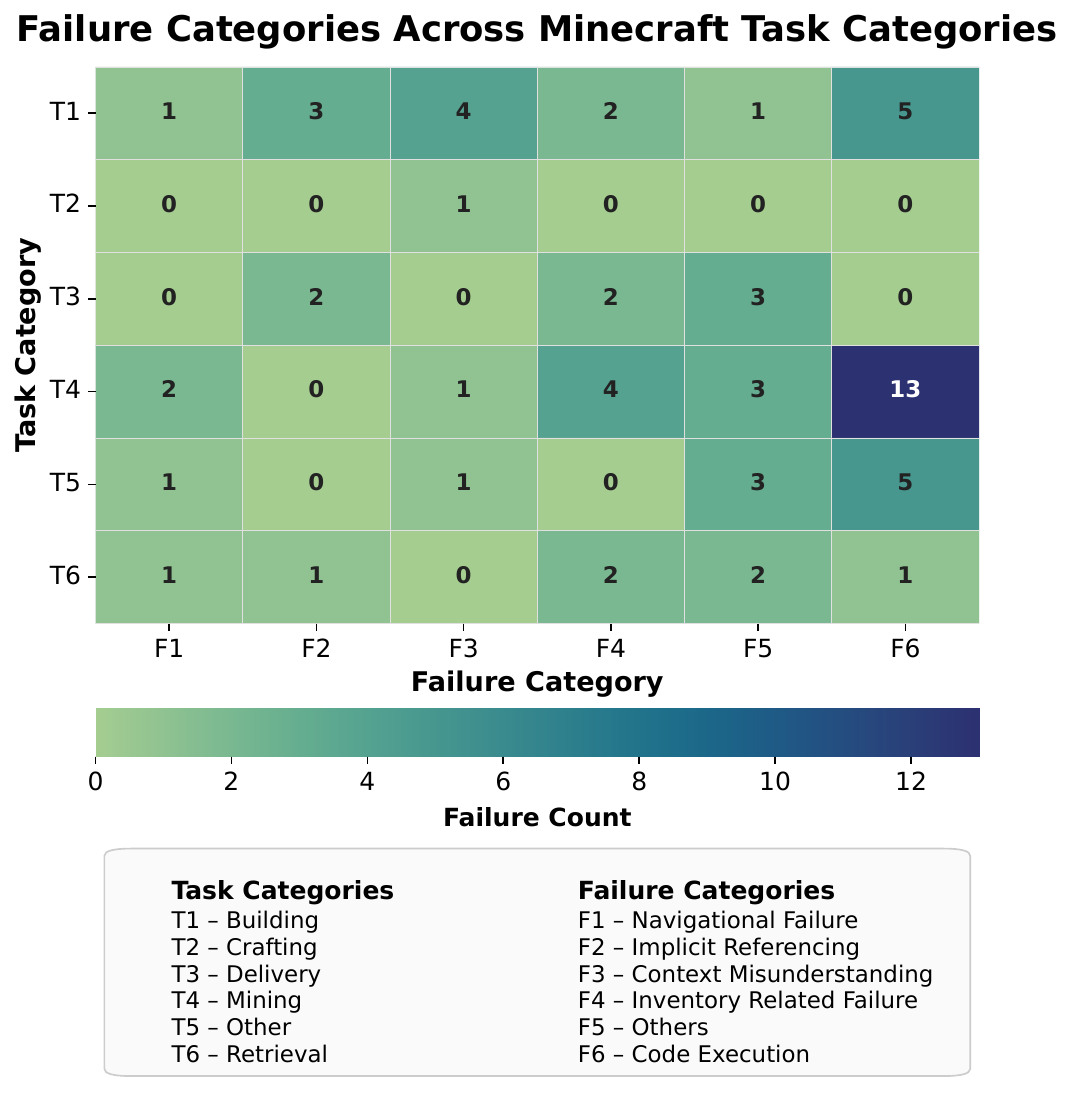}
  \caption{Cross-tabulation of failure categories by Minecraft task type. 
  Code/inventory issues cluster around mining and construction tasks, while 
  referencing failures and context misunderstandings appear more frequently 
  during retrieval and navigation.}
  \label{fig:task-failure-crossmap}
\end{figure}
\vspace{-12pt}


\paragraph{Setup.}
Each session included: (1) a brief interface walkthrough; (2) guided demonstrations covering landmark recall, retrieval, resource gathering with failure recovery, and contextual suggestion/tool awareness; (3) a participant-defined task; and (4) an exit survey. The harness logged routing and planning traces, clarifications, memory reads/writes, execution events, code-generation attempts, validator outputs, and synchronized screen/audio recordings.

\paragraph{Denominator.}
Across \textbf{216} subtasks attempted by \textbf{\NumParticipants{}}, we observed \textbf{71} subtask failures, yielding a \textbf{\SubtaskFailureRate{}} subtask-level failure rate. All results below reflect only these observed traces.

  


\HeroFull{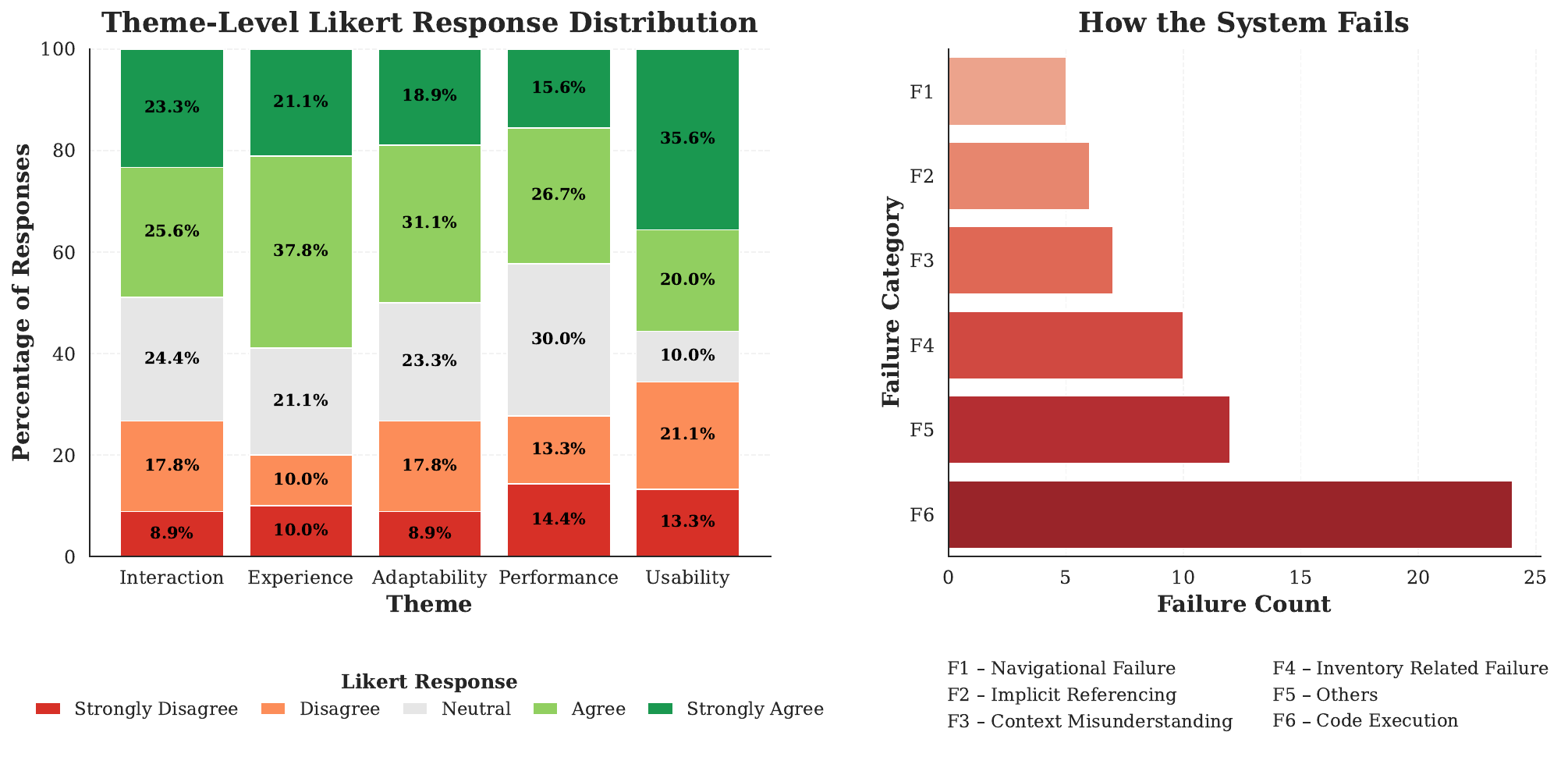}
{\textbf{Player experience and system breakdown patterns.}
  \emph{Left:} Likert-scale responses from participants across interaction quality, 
  memory utility, task performance, and UI usability.
  \emph{Right:} Distribution of breakdown types across 216 subtasks including 
  code/execution faults, inventory/tool misuse, referencing errors, context misunderstandings, 
  and navigation failures.
}
{fig:likert-failure}


\FloatBarrier


\subsection{Aggregate Engagement}
Across participants, \textbf{44} high-level tasks were attempted (\(M = 4.9\) per user), spanning building, mining, retrieval, navigation, and crafting. Despite breakdowns, most users completed goals after one or more attempts by revising commands or collaboratively troubleshooting with the interface.

\subsection{Perception and Interaction Quality}
Participants reported generally positive impressions of interaction quality and interface usability (Fig.~\ref{fig:likert-failure}). 

\textbf{Interaction and Communication} received agreement from 7/8 users, while both \textbf{Usability \& Overall Experience} exceeded 75\% agreement. Memory-based recall was cited as helpful by 6/8 participants, though several requested stronger persistence (e.g., “If I correct it once, it should remember next time.”).

\subsection{Breakdown Patterns and Recoverability}
The framework surfaces failures without assuming causality. Recurring patterns included:

\begin{itemize}[leftmargin=*,itemsep=2pt]
  \item \textbf{Code/execution failures} (\(n=24\)): invalid parameters (e.g., \texttt{Vec3}, misfired triggers, NaN resource counts).
  \item \textbf{Inventory/tool issues} (\(n=10\)): missing or misused tools (e.g., harvesting with a pickaxe); observed across 4 participants.
  \item \textbf{Context misunderstandings} (\(n=7\)): ambiguous or unclear semantics.
  \item \textbf{Referencing failures} (\(n=6\)): deictic phrasing such as “the block I'm looking at.”
  \item \textbf{Navigational failures} (\(n=5\)): shifting or ill-defined spatial targets.
\end{itemize}

\noindent
Task-type clustering patterns are summarized in Fig.~\ref{fig:task-failure-crossmap}. Participants often recovered from failures: \textbf{5/8} successfully completed a previously failed task by simplifying goals, clarifying parameters, or guiding execution manually (e.g., specifying coordinates or narrowing permissible action ranges).

\subsection{Transparency and Language Alignment}
Participants were more forgiving when failures were accompanied by contextual reasoning (e.g., “NaN cobblestone needed—could you clarify?”). Over time, players refined commands—adding coordinates, naming tools or furnaces, specifying variants—while the agent initiated targeted clarifications (“Which pickaxe?”). This co-adaptation produced a small but meaningful learning effect.

\subsection{Expectations of Companionship and Shared Perception}
Players frequently anthropomorphized the agent and expected shared visual grounding (“Look where I’m looking”). Several requested visibility into memory contents to bridge misunderstandings, suggesting that surfaced memory may meaningfully support trust and alignment.

\subsection{Design Implications}
\begin{itemize}[leftmargin=*,itemsep=2pt]
  \item \textbf{Transparent feedback fosters resilience.} Short explanations encouraged users to persist after breakdowns.
  \item \textbf{Language alignment is mutual.} Players refined phrasing; agents should reciprocate with clarifications and summaries.
  \item \textbf{Memory visibility builds trust.} Surfacing remembered facts reduces confusion about what the agent knows.
  \item \textbf{Recoverability matters more than perfection.} Bounded repair loops and partial replans are effective levers.
  \item \textbf{Social presence matters.} Egocentric camera cues and responsive behaviors strengthened the feeling of a companion agent.
\end{itemize}

\noindent
We report only the collected observations and denominator-aware outcomes. The framework is publicly released to enable reproducible evaluation on user-authored tasks under identical conditions.

\section{Conclusion}
We introduce \textsc{MineNPC-Task}, a practical benchmark for evaluating mixed-initiative, memory-aware LLM agents in \emph{Minecraft} using only public in-game interfaces. 

Tasks are elicited from expert co-play, normalized into compact templates with explicit preconditions, and paired with simple validators that judge completion from bounded, in-world evidence. 

An initial snapshot with GPT-4o over 44 tasks (216 subtasks) highlights where agents struggle slot clarification, preconditions, memory reuse, and code execution yielding a \(\sim\)33\% subtask failure rate. 

We hope this benchmark encourages models and methods that plan across dependent steps, ask when uncertain, and ground memory in observable state without relying on hidden shortcuts.

\section{Limitations}

\textbf{Scope.} Results apply to \emph{Minecraft} with a Mineflayer client under a bounded-knowledge policy. Portability to other engines or sensing/action envelopes is not evaluated.

\textbf{Model coverage.} We report a single-model snapshot (GPT-4o). There are no ablations or cross-model comparisons, so relative performance remains an open question.

\textbf{Task coverage.} The suite is expert-elicited and modest in size. It captures common goals we observed but does not exhaust the space of open-world play; selection bias is possible.

\textbf{Measurement granularity.} Validators return pass/fail from in-world traces. This improves reproducibility but under-represents partial progress and does not yet score cost or efficiency.

\textbf{Run-time variance.} Live co-play introduces practical variability (e.g., pathing, chunk loading). Policies reduce but do not eliminate noise; we report denominators to contextualize outcomes.

\textbf{Interaction design choices.} The framework prioritizes a single clarifying question and short plan previews. Richer multi-turn clarification or alternative planning styles may yield different results.

\textbf{Telemetry and auditability.} Released logs emphasize end-state evidence and high-level traces. Finer-grained diagnostics (e.g., pre-execution static checks) are limited and left to future iterations.

\textbf{Generalization of corrections.} While preferences and names are stored, persistent generalization of user corrections across sessions is limited in the current release.

\section{Future Work}

\textbf{Comparable baselines.} Add multi-model evaluations under identical perception–action contracts and policies, enabling controlled comparisons and public leaderboards.

\textbf{Expanded task pool.} Grow the expert-derived templates with parameterized variants and richer dependencies (e.g., multi-session builds), while keeping validators simple and reproducible.

\textbf{Richer metrics.} Report partial credit and efficiency (time, distance, resource cost), number of clarifications, and repair rate to complement pass/fail outcomes.

\textbf{Pre-execution checks and telemetry.} Introduce lightweight static checks for common API/parameter faults and expose finer-grained traces to improve diagnosis and replication.

\textbf{Targeted robustness probes.} Add focused tests for egocentric references and tool–affordance errors; surface memory contents and staleness in the UI to align expectations.

\textbf{Artifacts and reproducibility.} Continue releasing templates, validators, prompts, and logs so others can rerun the suite, contribute tasks, and extend the benchmark over time.

\bibliographystyle{ACM-Reference-Format}
\bibliography{agent}

\clearpage                       
\appendix
\FloatBarrier                    

\newenvironment{FullWidthListing}
{
  \begin{figure}[!htbp]
    \centering
    \begin{minipage}{0.97\linewidth}
    \vspace{4pt}
}
{
    \end{minipage}
  \end{figure}
  \FloatBarrier
}


\AppendixOneColumn

\FloatBarrier

\captionsetup[figure]{skip=8pt}
\captionsetup[table]{skip=6pt}
\setlength{\textfloatsep}{12pt plus 2pt minus 2pt}
\setlength{\floatsep}{12pt plus 2pt minus 2pt}
\setlength{\intextsep}{12pt plus 2pt minus 2pt}

\tcbset{
  enhanced jigsaw,
  breakable,
  colback=gray!2,
  colframe=black!20,
  boxrule=0.4pt,
  left=6pt,right=6pt,top=6pt,bottom=6pt,
  before skip=10pt,
  after skip=14pt
}

\appendix
\onecolumn
\begin{center}
{\Large \textbf{APPENDIX}} \\[6pt]
\end{center}

\section{User Interface and Runtime Architecture}
\label{app:interface}

\vspace{6pt}

\subsection{Player-Facing Interface Design}

\paragraph{Design Philosophy.}
Our UI emphasizes uninterrupted gameplay: player control remains primary. 
The right-hand chat rail presents all agent interactions, including 
\texttt{PlanPreview}, clarifying queries, live execution updates, and validator feedback. 
These elements surface inline and incrementally, avoiding disruptive modals. 
All interactions are timestamped and logged consistent with the 
Mineflayer-only evaluation policy. A video demonstration of this interaction flow is available at
\href{https://youtu.be/7hqQl9XhjAk}{\texttt{link}}.

\vspace{10pt}

\begin{figure}[!hbt]
  \centering
  \includegraphics[width=\columnwidth]{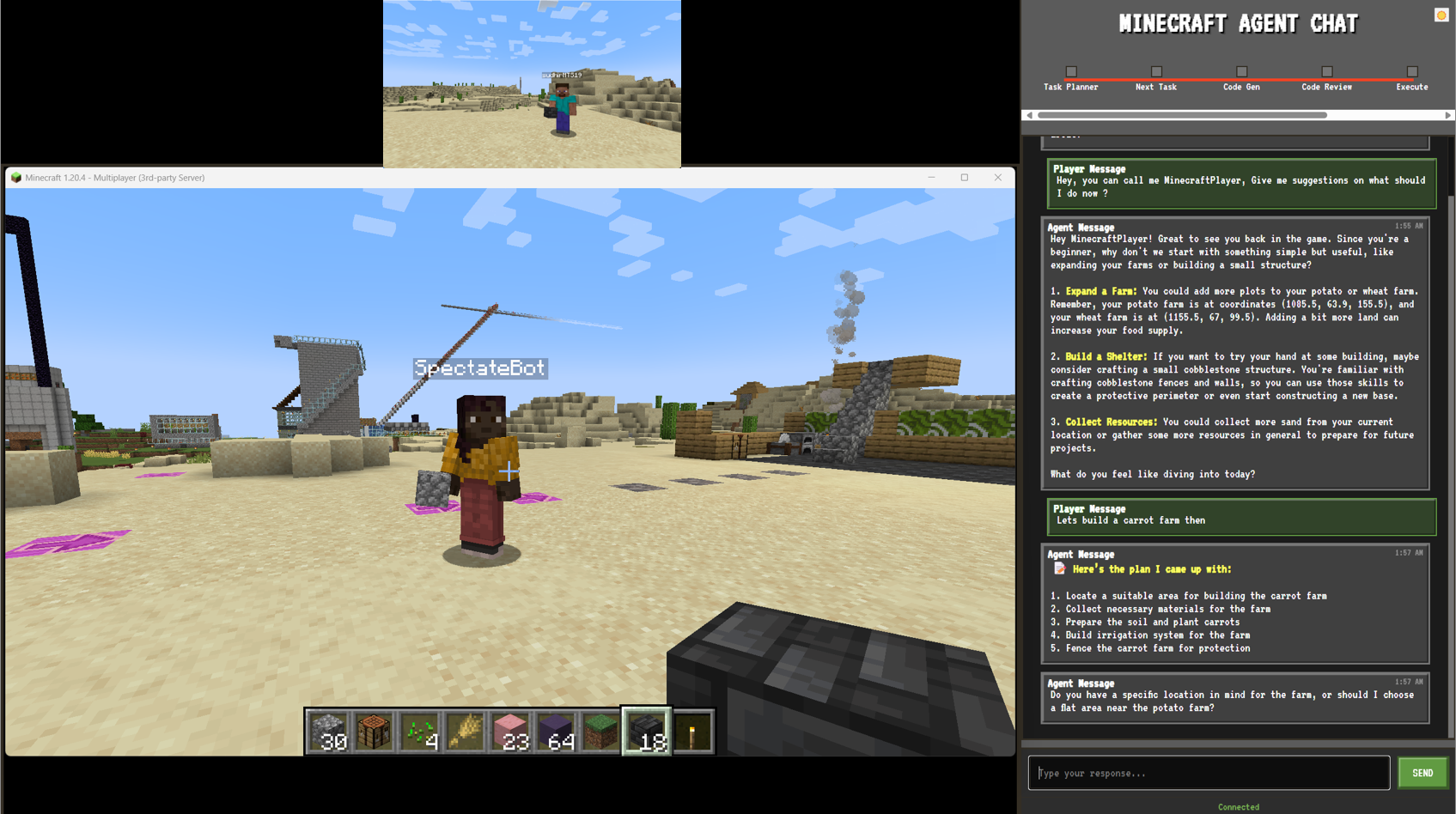}
  \caption{\textbf{Player interface for memory-grounded agent interaction.}
  \emph{Left:} Unobstructed main viewport with minimal spectator NPC.
  \emph{Right:} Lightweight side-rail chat for agent I/O plan previews, 
  clarification, validator summaries, and progress toasts.}
  \label{fig:ui-appx}
\end{figure}

\vspace{10pt}

\medskip

\subsection{Core Runtime Data Structures}
Below we document core runtime schemas used by the planner, memory, and validator modules. Each schema is designed for interpretability, serialization, and traceability across the pipeline.

\FloatBarrier
\vspace{6pt}

\vspace{2pt}
\FloatBarrier

\begin{table}[!htbp]
  \centering
  \caption{\texttt{BotState}: snapshot of world and agent state at a given tick.}
  \label{tab:botstate}
  \begin{tabularx}{\linewidth}{@{} l l X @{}}
    \toprule
    \textbf{Field} & \textbf{Type} & \textbf{Description} \\
    \midrule
    time & int & In-game tick count when captured. \\
    position & Position & Absolute world coordinates \((x,y,z)\). \\
    health, hunger & int & Current health and hunger values. \\
    nearby\_blocks & List\textless{}str\textgreater{} & Surrounding block types in range. \\
    equipment & List\textless{}str\textgreater{} & Equipped items (tools, armor). \\
    inventory & Inventory & Complete item inventory with counts. \\
    chests & List\textless{}str\textgreater{} & Recently seen or accessed chest locations. \\
    movement & Movement & Current movement goal and state. \\
    nearby\_entities & List\textless{}Entity\textgreater{} & Visible entities with metadata. \\
    recent\_chat & List\textless{}str\textgreater{} & Recent messages in local chat. \\
    active\_effects & List\textless{}str\textgreater{} & Active potion/status effects. \\
    environment & Environment & Environmental context (biome, weather, time). \\
    target\_block & Optional\textless{}str\textgreater{} & Current interaction target (if any). \\
    \bottomrule
  \end{tabularx}
\end{table}

\FloatBarrier
\vspace{8pt}

\vspace{2pt}
\FloatBarrier

\begin{table}[!htbp]
  \centering
  \caption{\texttt{TaskModel}: normalized subtask representation.}
  \label{tab:taskmodel}
  \begin{tabularx}{\linewidth}{@{} l l X @{}}
    \toprule
    \textbf{Field} & \textbf{Type} & \textbf{Description} \\
    \midrule
    index & int & Unique index within plan. \\
    subtask & str & Natural language description. \\
    dependencies & List\textless{}int\textgreater{} & Indices of prerequisite subtasks. \\
    parameters & Any & Arguments required for execution. \\
    status & str & Execution state: \texttt{pending} \(|\) \texttt{in-progress} \(|\) \texttt{completed}. \\
    clarifying\_questions & List\textless{}str\textgreater{} & Outstanding disambiguation questions. \\
    success\_criteria & str & Machine-checkable completion condition. \\
    \bottomrule
  \end{tabularx}
\end{table}

\FloatBarrier
\vspace{8pt}

\vspace{2pt}
\FloatBarrier

\begin{table}[!htbp]
  \centering
  \caption{\texttt{ValidationOutput}: task evaluation with reasoning.}
  \label{tab:validation}
  \begin{tabularx}{\linewidth}{@{} l l X @{}}
    \toprule
    \textbf{Field} & \textbf{Type} & \textbf{Description} \\
    \midrule
    success, failure & Optional\textless{}str\textgreater{} & Human-readable explanation. \\
    state\_changes & List\textless{}StateChange\textgreater{} & Key in-world deltas. \\
    chat\_insights & List\textless{}ChatMessage\textgreater{} & Relevant parsed chat excerpts. \\
    suggestions & Optional\textless{}str\textgreater{} & Repair/retry recommendations. \\
    question\_for\_player & Optional\textless{}str\textgreater{} & Clarification prompt if ambiguous. \\
    is\_task\_completed & bool & Binary task verdict. \\
    final\_result & Optional\textless{}str\textgreater{} & Validator's reasoning summary. \\
    confidence\_score & int & Certainty level \([0,10]\). \\
    \bottomrule
  \end{tabularx}
\end{table}

\FloatBarrier
\vspace{10pt}

\newpage
\subsection{Pydantic Schema Definitions}
\label{app:pydantic}
\vspace{4pt}
\FloatBarrier

\begin{FullWidthListing}
\caption{Production runtime schemas in Pydantic.}
\label{lst:schemas}
\begin{lstlisting}[language=Python]
class BotState(BaseModel):
    """Agent state snapshot"""
    time: int
    position: Position
    health: int
    hunger: int
    nearby_blocks: List[str]
    equipment: List[str]
    inventory: Inventory
    chests: List[str]
    movement: Movement
    nearby_entities: List[Entity]
    recent_chat: List[str]
    active_effects: List[str]
    environment: Environment
    target_block: Optional[str]

class TaskModel(BaseModel):
    """Structured subtask with dependencies"""
    index: int
    subtask: str
    dependencies: List[int] = []
    parameters: Any = {}
    status: str = "pending"
    clarifying_questions: List[str] = []
    success_criteria: str

class ValidationOutput(BaseModel):  
    """Validator outputs"""
    success: Optional[str]
    failure: Optional[str]
    state_changes: List[StateChange] = []
    chat_insights: List[ChatMessage] = []
    suggestions: Optional[str]
    question_for_player: Optional[str]
    is_task_completed: bool
    final_result: Optional[str]
    confidence_score: int
\end{lstlisting}
\end{FullWidthListing}

\clearpage

\section{Task Planning and Replanning Prompts}
\label{app:planning}

\vspace{10pt}

\subsection{Initial Task Planning}

\vspace{10pt}

\subsubsection{System Prompt: Initial Task Planner}

\begin{tcolorbox}[title=System Prompt: Initial Task Planner]
You are an advanced \textbf{Minecraft AI Task Planner}, responsible for generating structured and executable task plans for a personalized Minecraft assistant. You use the \textbf{Mineflayer API} to ensure tasks are well-defined, achievable, and optimized based on real-time game conditions and user preferences.

\textbf{Core Responsibilities:}
\begin{itemize}[leftmargin=*]
  \item Break down high-level user tasks into structured, sequential subtasks.
  \item Minimize unnecessary clarifications via bot state and user context.
  \item Generate plans that are executable within current game conditions.
  \item \textbf{Never} modify previously completed tasks during replanning.
\end{itemize}
\end{tcolorbox}

\begin{lstlisting}[style=json,caption={Expected JSON output for the initial task planner.}]
{
  "task_plan": [
    {
      "index": 1,
      "subtask": "Clear, actionable description",
      "dependencies": [],
      "parameters": {},
      "status": "pending",
      "clarifying_questions": [],
      "success_criteria": "Measurable completion condition"
    }
  ]
}
\end{lstlisting}

\section{Clarification and Feedback Management}
\label{app:clarification}

\vspace{10pt}

\subsection{Subtask Clarification System}

\vspace{10pt}

\begin{tcolorbox}[title=System Prompt: Subtask Clarification]
Given a task with outstanding clarifying questions in JSON, generate the missing information while preserving all existing fields.
\end{tcolorbox}

\begin{lstlisting}[style=json,caption={Example task requiring clarification.}]
{
  "index": 2,
  "subtask": "Retrieve a sword from the chest",
  "parameters": {"item_name": "sword"},
  "clarifying_questions": ["What type of sword do you need?"]
}
\end{lstlisting}

\begin{lstlisting}[style=json,caption={Clarification response example.}]
[
  {
    "question": "What type of sword do you need?",
    "answer": "diamond_sword"
  }
]
\end{lstlisting}

\section{Validation, Code Generation, and System Updates}
\label{app:validation}

\vspace{10pt}

\subsection{Subtask Validation System}

\begin{tcolorbox}[title=System Prompt: Subtask Validator]
You are a \textbf{Validation AI Agent} responsible for evaluating a Minecraft bot's task execution via world-state analysis and producing structured reports.
\end{tcolorbox}

\subsection{Player Landmark Tracking System}

\begin{lstlisting}[style=json,caption={Landmark update schema.}]
{
  "landmarks": {
    "weapon storage": {
      "coordinates": [x, y, z],
      "radius": 6
    }
  }
}
\end{lstlisting}

\subsection{Code Generation and Review Examples}

\begin{lstlisting}[style=json,caption={Code generation output example.}]
{
  "code": "async function task(bot) { /* implementation */ }"
}
\end{lstlisting}

\begin{lstlisting}[style=json,caption={Code review output example.}]
{
  "valid": true,
  "confidence": 0.9,
  "comments": [
    {"line": 5,  "comment": "Use mineBlock() instead of bot.dig()"},
    {"line": 12, "comment": "Add null check for bot.inventory"}
  ]
}
\end{lstlisting}

\clearpage

\providecommand{\appxsep}{\vspace{0.8\baselineskip}}
\providecommand{\appxsmallsep}{\vspace{0.45\baselineskip}}

\setlist[itemize]{leftmargin=*, topsep=4pt, itemsep=2pt}
\setlist[enumerate]{leftmargin=*, topsep=4pt, itemsep=2pt}

\tcbset{
  appxbox/.style={
    colback=gray!2, colframe=black!15,
    enhanced, sharp corners,
    boxsep=1ex, left=1.2ex, right=1.2ex, top=0.8ex, bottom=0.8ex,
    before skip=12pt, after skip=14pt, breakable
  }
}

\BackToTwoColumn

\section{\textsc{MineNPC-Task} and Subtask Breakdown}
\label{app:tasksuite}

\appxsmallsep
The following \textbf{44 tasks} constitute the \emph{complete} \textsc{MineNPC-Task} suite used in our study. They were elicited from expert co-play rather than synthesized prompts, span multiple gameplay domains, and define the benchmark workload for evaluating agents under our bounded-knowledge policy.

\appxsep
\subsection{High-Level Task Categories in \textsc{MineNPC-Task}}

\subsubsection{Resource Collection and Mining}
\begin{itemize}
  \item Mine and collect cobblestones and deliver them.
  \item Mine cobblestones starting from specific coordinates (1145, 58, 56).
  \item Use pickaxe to mine 20 blocks of cobblestone in front of player.
  \item Use golden pickaxe to mine 5 cobblestone blocks.
  \item Mine coal nearby using diamond pickaxe.
  \item Continue mining 4 coal blocks, then deliver cobblestone and coal.
  \item Mine 16 stones from designated spot, find and mine 4 coal blocks.
  \item Mine iron ore for the player.
  \item Take iron pickaxe and get at least 10 coal blocks.
  \item Drop 5 cobblestones in front of player.
  \item Give 4 cobblestones to player.
  \item Harvest oak logs (at least 20 blocks) using an axe.
\end{itemize}

\appxsmallsep
\subsubsection{Tool and Equipment Management}
\begin{itemize}
  \item Pick up a pickaxe from the pickaxe chest.
  \item Get a pickaxe from the chest at the best entrance.
  \item Get a pickaxe from the chest at the mine.
  \item Grab an iron pickaxe from the chest at \texttt{ChestEntrance}.
  \item Bring iron pickaxe from \texttt{iron\_pickaxe\_chest}.
  \item Get iron pickaxe from ``Storage''.
  \item Come to player and drop the coal.
  \item Give cobblestone that was mined.
\end{itemize}

\appxsmallsep
\subsubsection{Agriculture and Food}
\begin{itemize}
  \item Harvest 5 wheat.
  \item Harvest wheat and craft bread.
  \item Collect wheat and make 3 pieces of bread.
  \item Harvest wheat and craft 3 pieces of bread if needed.
  \item Collect 18 wheat blocks.
  \item Help plant more seeds.
  \item Go to ``Farm'' to harvest wheat for 2 bread.
  \item Walk around and pick up all harvested wheat.
\end{itemize}

\appxsmallsep
\subsubsection{Construction and Building}
\begin{itemize}
  \item Build a pyramid from sand (5$\times$5 base).
  \item Build a simple house with roof at \texttt{build\_spot}.
  \item Collect dirt and build a dirt house.
  \item Build walls for the house.
  \item Build a wall 10 blocks long and 2 blocks high next to house.
  \item Make a 4$\times$2 vertical wall at specified coordinates.
  \item Decorate house with blue stained glass at specific coordinates.
\end{itemize}

\appxsmallsep
\subsubsection{Crafting and Processing}
\begin{itemize}
  \item Turn stripped oak logs into planks, then into stairs.
  \item Come to location and craft stairs.
  \item Collect materials from preferred building list.
\end{itemize}

\appxsmallsep
\subsubsection{Storage and Inventory}
\begin{itemize}
  \item Put wood-related blocks into the chest in front of player.
  \item Get stained glass from design chest (10 blocks).
  \item Give 5 pieces of magenta stained glass and 2 pieces of redstone block.
  \item Bring 25 blocks of stripped cherry wood and 4 lanterns.
  \item Give all cherry wood and lanterns.
  \item Drop brain coral block, exposed copper, and terracotta.
\end{itemize}

\FloatBarrier
\appxsep
\subsection{Full Subtask Decomposition for \textsc{MineNPC-Task}}

\begin{small}
\begin{enumerate}
  \item Mine cobblestone.
  \item Mine cobblestone starting from specified coordinates.
  \item Navigate to the nearest chest.
  \item Locate and verify crafting table at \texttt{wheat\_farm}.
  \item Locate the nearest lava source.
  \item Walk around to find a nearby crafting table.
  \item Locate a nearby tree to collect wood.
  \item Navigate to the cave or mine shaft.
  \item Move to the chest in front of the player.
  \item Move to the chest location.
  \item Locate player's boat nearby.
  \item Return to user with collected items.
  \item Navigate to the Pickaxe chest.
  \item Go to the starting coordinate for mining.
  \item Find the nearest chest.
  \item Navigate to the coal site.
  \item Go to player \texttt{ThorThunder92}.
  \item Locate the player to give wheat.
  \item Navigate to the closest coal ore block.
  \item Travel to \texttt{build\_spot}.
  \item Navigate to \texttt{ChestEntrance}.
  \item Navigate to the \texttt{wheatFarm} location.
  \item Navigate to a dirt-rich area.
  \item Navigate to the Storage location.
  \item Locate a cave or mine shaft.
  \item Travel to Farm.
  \item Move around the farm area to locate dropped wheat.
  \item Go to the farm area.
  \item Approach the nearest player.
  \item Locate oak trees in the surrounding area.
  \item Move to the nearest oak tree.
  \item Return to the original position or safe location.
  \item Locate the user.
  \item Navigate to the treasure box location.
  \item Navigate to the wood storage area.
  \item Go to the mine.
  \item Locate the nearest cobblestone block.
  \item Move to the cobblestone block.
  \item Locate the nearest player to hand over cobblestone.
  \item Move to the nearest player.
  \item Navigate to the wheat farm.
  \item Locate player ``ThorThunder92''.
  \item Go to the design storage chest.
  \item Locate the design storage chest.
  \item Locate the block storage chest.
  \item Navigate to utilities area.
  \item Go to the block storage.
  \item Go to the chest in the utilities.
  \item Move to specified coordinates.
  \item Go to the chest location.
  \item Harvest 5 wheat.
  \item Mine 16 cobblestone.
  \item Collect nearby wheat items.
  \item Search for and pick up an axe.
  \item Collect stripped cherry wood.
  \item Collect stripped cherry wood from wood storage.
  \item Gather 5 pieces of glass.
  \item Collect 10 blocks of stained glass from the chest.
  \item Collect 2 pieces of redstone block from block storage.
  \item Collect additional wheat.
  \item Collect dropped wheat.
  \item Collect material for building walls.
  \item Collect 32 terracotta blocks from chest.
  \item Collect enough building material for the wall.
  \item Explore to find cobblestone.
  \item Mine cobblestone until 20 are collected.
  \item Collect wheat from the ripened wheat farm.
  \item Collect sufficient sand for pyramid construction.
  \item Gather required materials for pyramid.
  \item Get 18 items of wheat from wheat crops.
  \item Harvest matured wheat.
  \item Mine the coal ore block.
  \item Search for iron ore.
  \item Harvest wheat from the farm.
  \item Harvest oak logs using stone axe.
  \item Collect dirt blocks.
  \item Pick up the golden pickaxe from treasure box.
  \item Collect 25 blocks of stripped cherry wood.
  \item Collect 4 lanterns.
  \item Collect 5 pieces of magenta stained glass.
  \item Mine 3 blocks of coal from identified deposits.
  \item Mine 5 cobblestone blocks using golden pickaxe.
  \item Check for mature wheat and harvest.
  \item Retrieve 64 green glazed terracotta from chest.
  \item Locate and open the chest.
  \item Open the chest and collect 32 terracotta blocks.
  \item Open the Pickaxe chest.
  \item Open the chest at \texttt{ChestEntrance}.
  \item Open the treasure box.
  \item Open the design storage chest.
  \item Check wood storage for stripped cherry wood.
  \item Check style storage for lanterns.
  \item Collect brain coral block from sea or storage.
  \item Gather exposed copper from storage or crafting.
  \item Gather magenta stained glass from storage.
  \item Check the chest for blue or green colored blocks.
  \item Check for blue terracotta block.
  \item Craft 2 breads from harvested wheat.
  \item Craft bread.
  \item Use the nearby crafting table to craft bread.
  \item Craft planks from available logs.
  \item Craft a stone axe using available material.
  \item Craft bread (from farm wheat).
  \item Craft oak planks from stripped oak logs.
  \item Craft bread using wheat.
  \item Craft 3 bread using collected wheat.
  \item Load sand into the located furnace.
  \item Add fuel to the furnace to start smelting.
  \item Retrieve 3 glass blocks from the furnace.
  \item Locate a nearby furnace.
  \item Identify the furnace location in front of the bot.
  \item Craft planks from available logs.
  \item Check the inventory for oak logs.
  \item Ensure crafting table is properly placed or accessible.
  \item Locate the seed dropped by the user.
  \item Plant wheat seeds within a radius of 10 blocks.
  \item Identify all farmland blocks within 10-block radius.
  \item Plant wheat seeds on all identified farmland blocks.
  \item Locate seeds in the inventory.
  \item Select building material from inventory.
  \item Select diamond pickaxe from inventory.
  \item Retrieve \texttt{blue\_stained\_glass} from inventory.
  \item Verify iron pickaxe in the inventory.
  \item Check for inventory.
  \item Select 4 cobblestones from inventory.
  \item Retrieve cobblestone from inventory.
  \item Check inventory for diamond pickaxe.
  \item List all items from inventory.
  \item Locate golden pickaxe in inventory or known chest.
  \item Identify wood-related items in the inventory.
  \item Build 4 by 2 blocks.
  \item Select current location as pyramid build location.
  \item Select suitable location to build the pyramid.
  \item Build base layer of pyramid with sand.
  \item Lay the foundation layer of 5$\times$5.
  \item Lay the second layer of 4$\times$4.
  \item Lay the third layer of 3$\times$3.
  \item Lay the fourth layer of 2$\times$2.
  \item Place the final block at the top.
  \item Begin constructing the house framework using planks.
  \item Construct the walls of the dirt house.
  \item Find a suitable location for building the dirt house.
  \item Locate house to start decorating.
  \item Place \texttt{blue\_stained\_glass} at specified coordinates.
  \item Place blocks to ascend to the surface.
  \item Determine location of the house.
  \item Return to X and transfer coal.
  \item Transfer 4 cobblestones to the nearest player.
  \item Retrieve a pickaxe from the chest.
  \item Give the cobblestone to the player.
  \item Return to the player and give the bread.
  \item Give 5 cobblestones to the closest player.
  \item Return to the user and give cobblestones.
  \item Transfer coal to the player \texttt{ThorThunder92}.
  \item Bring an iron pickaxe from \texttt{iron\_pickaxe\_chest}.
  \item Return to the player and present the iron pickaxe.
  \item Transfer coal and cobblestone to Rene.
  \item Adjust pathfinding to get close to X and deliver iron pickaxe.
  \item Return to the user and deliver the iron pickaxe.
  \item Punch the player.
  \item Drop oak stair in front of the bot.
  \item Drop the terracotta.
  \item Drop the brain coral block.
  \item Drop the exposed copper.
  \item Drop 5 cobblestones in front of the player.
  \item Deliver the 3 bread to the player.
  \item Give all cherry wood and lanterns to the player.
  \item Retrieve the gold pickaxe from the chest.
  \item Bring the golden pickaxe to the user.
  \item Transfer the 64 green glazed terracotta to player.
  \item Locate the chest containing the stone pickaxe.
  \item Retrieve a pickaxe from the Pickaxe chest.
  \item Retrieve an iron pickaxe from the chest.
  \item Locate the nearest chest.
  \item Retrieve diamond pickaxe from utilities area chest.
  \item Check if wheat farm has sufficient mature wheat.
  \item Check the maturity of wheat crops at ripened wheat farm.
  \item Move to the starting point for wheat collection.
  \item Align with mature wheat crop.
  \item Ensure the farming tool is functional.
  \item Locate a sand collection area.
  \item Navigate to pyramid location.
  \item Look around and identify all visible coal ore blocks.
  \item Attempt to harvest additional wheat.
  \item Identify matured wheat crops in nearby farmland.
  \item Locate and open the chest.
  \item Verify if the chest at the mine contains an iron pickaxe.
  \item Retrieve an iron pickaxe from the chest.
  \item Return to the player and give the iron pickaxe.
  \item Use iron pickaxe to mine 20 cobblestone blocks.
  \item Check for lime terracotta block.
  \item Retrieve the diamond pickaxe from the chest.
  \item Locate nearby coal deposits.
  \item Navigate to a mining area where cobblestone can be found.
  \item Locate the closest player entity.
  \item Go to the wheat farm to harvest wheat.
  \item Go to the location of the house.
  \item Go to the storage chest containing terracotta blocks.
  \item Locate the green glazed terracotta in design storage chest.
  \item Locate player to deliver the item.
  \item Move to the designated wall construction area next to the house.
  \item Gather necessary building materials.
  \item Locate and verify presence of crafting table.
  \item Locate a nearby tree to collect wood.
  \item Locate 'userName'.
  \item Harvest additional wheat if needed.
  \item Identify all farmland blocks within radius.
  \item Navigate to closest coal ore block.
  \item Locate cobblestone.
  \item Navigate to cave or mine shaft.
  \item Return to user with collected items.
  \item Move to chest in front of player.
  \item Navigate to wood storage area.
  \item Check style storage for lanterns.
  \item Navigate to utilities area.
  \item Move to specified coordinates.
\end{enumerate}
\end{small}

\appxsep
\subsection{Complexity Analysis for \textsc{MineNPC-Task}}

\paragraph{Suite Statistics.}
\begin{itemize}
  \item \textbf{Total \textsc{MineNPC-Task} Tasks:} 44 distinct user requests.
  \item \textbf{Total Subtasks Generated:} 216 atomic operations.
  \item \textbf{Average Subtasks per Task:} 4.9 steps.
  \item \textbf{Task Categories:} 6 major functional domains.
  \item \textbf{Complexity Range:} 1--12 subtasks per high-level task.
\end{itemize}

\appxsmallsep
\paragraph{Domain Distribution (within the suite).}
\begin{itemize}
  \item \textbf{Resource Collection:} 32\% of subtasks (mining, harvesting, gathering).
  \item \textbf{Navigation and Movement:} 28\% (pathfinding, location-based).
  \item \textbf{Item Management:} 18\% (transfer, storage, inventory).
  \item \textbf{Construction:} 12\% (building, placement, crafting).
  \item \textbf{Tool Management:} 6\% (retrieval, selection, delivery).
  \item \textbf{Interaction:} 4\% (player communication, coordination).
\end{itemize}

\appxsmallsep
\paragraph{System Requirements Demonstrated by \textsc{MineNPC-Task}.}
\begin{itemize}
  \item \textbf{Spatial Reasoning:} Coordinate-based navigation and construction.
  \item \textbf{Resource Management:} Inventory tracking and optimization.
  \item \textbf{Multi-Step Planning:} Complex task decomposition and sequencing.
  \item \textbf{Player Coordination:} Item delivery and collaborative construction.
  \item \textbf{Environmental Awareness:} Dynamic world state adaptation.
  \item \textbf{Tool Specialization:} Context-appropriate tool selection and usage.
\end{itemize}

\FloatBarrier
\appxsep


\section{User Personalization and Context Management}
\label{app:usercontext}

\appxsmallsep
\noindent
\begin{tcolorbox}[appxbox,
  title=Structured User Data Format (\texttt{user\_data.json}), fonttitle=\bfseries]
This file maintains comprehensive personalized information about the player, continuously gathered through conversation, task outcomes, and memory updates. The structure evolves to reflect player preferences and behaviors.
\medskip

\textbf{Base Structure Example:}
\begin{lstlisting}[language=json, basicstyle=\ttfamily\scriptsize]
{
  "user_info": {
    "user_name": "msrPlayer",
    "preferred_name": "",
    "experience_level": "Beginner",
    "preferred_language": "English"
  },
  "preferences": {
    "favorite_weapons": [],
    "specific_log_collection_location": [],
    "prefers_bot_assistance": null,
    "preferred_building_materials": [],
    "typical_play_style": ""
  },
  "world_knowledge": {
    "chest_locations": {},
    "safe_zones": [],
    "known_allies": [],
    "visited_locations": [],
    "landmarks": {},
    "resource_hotspots": []
  },
  "behavioral_insights": {
    "average_session_length": "",
    "interaction_frequency": "",
    "reaction_to_bot_failures": "",
    "preferred_communication_style": "",
    "task_completion_patterns": ""
  },
  "last_updated": ""
}
\end{lstlisting}
\end{tcolorbox}

\appxsmallsep
\noindent
\begin{tcolorbox}[appxbox, title=Dynamic Evolution Characteristics, fonttitle=\bfseries]
\begin{itemize}
  \item \textbf{Automatic Updates:} Refined via gameplay interactions and LLM inference.
  \item \textbf{Multi-Source Integration:} Combines chat logs, world observations, and action traces.
  \item \textbf{Behavioral Modeling:} Tracks player preferences, strategies, and reaction patterns.
  \item \textbf{Spatial Memory:} Expands with landmarks, resources, and zones of interest.
\end{itemize}
\end{tcolorbox}

\noindent
\begin{tcolorbox}[appxbox, title=System-Wide Usage, fonttitle=\bfseries]
\begin{itemize}
  \item \textbf{Planning Module:} Prioritizes tasks based on user preferences and play style.
  \item \textbf{Validation Module:} Interprets outcomes in light of behavioral patterns.
  \item \textbf{Code Generation:} Adapts implementations to experience level and materials.
  \item \textbf{Conversational Assistant:} Tailors tone, detail, and phrasing to the profile.
\end{itemize}
\end{tcolorbox}

\noindent
\begin{tcolorbox}[appxbox, title=Privacy and Persistence, fonttitle=\bfseries]
\begin{itemize}
  \item All profile data remains local to the player’s system.
  \item Enables long-term personalization across sessions.
  \item Updates are transparent and can be inspected via natural dialogue.
\end{itemize}
\end{tcolorbox}

\FloatBarrier

\BackToTwoColumn

\end{document}